\pdfoutput=1

\documentclass[11pt,twoside]{article}
\usepackage[letterpaper,inner=3.5cm,outer=2.5cm,top=2.5cm,bottom=2.5cm,pdftex]{geometry}
\usepackage[T1]{fontenc}
\usepackage[utf8]{inputenc}
\usepackage{lmodern}
\usepackage[pdftex]{graphicx}
\usepackage[pdftex]{hyperref}
\usepackage{amsfonts}
\usepackage{fancyhdr}
\usepackage{amsmath}
\usepackage{amsthm}
\usepackage{authblk}

\usepackage{mathtools}

\usepackage{algorithm}
\usepackage{algpseudocode}

\newcommand{\Rset}{\mathbb{R}}

\newcommand{\Zset}{\mathbb{Z}}

\newcommand{\Xset}{\mathcal{X}}
\newcommand{\Yset}{\mathcal{Y}}

\newcommand{\setA}{\mathcal{A}}
\newcommand{\setAp}{\mathcal{A}'}
\newcommand{\Prob}{P}
\newcommand{\Alg}{\Pi}
\newcommand{\ProbIn}{w}
\newcommand{\ProbSpace}{\mathcal{W}}

\newcommand{\lpopt}[1]{{{#1}^{*}}}

\newtheorem*{theorem*}{Theorem}
\newtheorem{theorem}{Theorem}

\newtheorem{mydef}{Definition}

\newcommand{\poly}{\mathrm{poly}}
\newcommand{\hull}{\mathrm{conv}}

\newcommand{\norm}[1]{||#1||}
\newcommand{\inner}[2]{\langle #1 , #2 \rangle}

\newcommand{\BigOh}[1]{O\left({#1}\right)}

\newcommand{\mute}[1]{}

\title{Differentiable Combinatorial Losses \\through Generalized Gradients of Linear Programs}

\author{Xi Gao\thanks{e-mail: gaox2@vcu.edu}}
\author{Han Zhang\thanks{e-mail: zhangh4@vcu.edu}}
\author{Aliakbar Panahi\thanks{e-mail: panahia@vcu.edu}}
\author{Tom Arodz\thanks{Corresponding author. e-mail: tarodz@vcu.edu}}
\affil{\mbox{Department of Computer Science}, \mbox{Virginia Commonwealth University} \mbox{Richmond, VA, USA}}
\date{}

\begin{document}
	\sloppy
	\maketitle
	
\begin{abstract}
Combinatorial problems with linear objective function play a central role in many computer science applications, and efficient algorithms for solving them are well known. However, the solutions to these problems are not differentiable with respect to the parameters specifying the problem instance – for example, shortest distance between two nodes in a graph is not a differentiable function of graph edge weights. Recently, attempts to integrate combinatorial and, more broadly, convex optimization solvers into gradient-trained models resulted in several approaches for differentiating over the solution vector to the optimization problem. However, in many cases, the interest is in differentiating over only the objective value, not the solution vector, and using existing approaches introduces unnecessary overhead. Here, we show how to perform gradient descent directly over the objective value of the solution to combinatorial problems. We demonstrate advantage of the approach in examples involving sequence-to-sequence modeling using differentiable encoder-decoder architecture with softmax or Gumbel-softmax, and in weakly supervised learning involving a convolutional, residual feed-forward network for image classification.
\end{abstract}

\section{Introduction}

Combinatorial optimization problems, such as shortest path in a weighted directed graph, minimum spanning tree in a weighted undirected graph, or optimal assignment of tasks to workers, play a central role in many computer science applications. We have highly refined, efficient algorithms for solving these fundamental problems \cite{Cormen, schrijver2003combinatorial}. However,  while we can easily find, for example, the minimal spanning tree in a graph, the total weight of the tree as function of graph edge weights is not differentiable. This problem hinders using solutions to  combinatorial problems as criteria in training models that rely on differentiability of the objective function with respect to the model parameters.

Losses that are defined by objective value of some feasible solution to a combinatorial problem, not the optimal one, have been recently proposed for  image segmentation using deep models \cite{zheng2015conditional,lin2016scribblesup}. These focus on a problem where some pixels in the image have segmentation labels, and the goal is to train a convolutional network that predicts segmentation labels for all pixels. For pixels with labels, a classification loss can be used. For the remaining pixels, a criterion based on a combinatorial problem -- for example the maximum flow / minimal cut problem in a regular, lattice graph connecting all pixels \cite{boykov2001fast} or derived, higher-level super-pixels \cite{lin2016scribblesup} -- is often used as a loss, in an iterative process of improving discrete segmentation labels \cite{zheng2015conditional,marin2019beyond}. In this approach, the instance of the combinatorial problem is either fixed, or depends only on the input to the network; for example, similarity of neighboring pixel colors defines edge weights. The output of the neural network gives rise to a feasible, but rarely optimal, solution to that fixed instance a combinatorial problem, and its quality is used as a loss. For example, pixel labeling proposed by the network is interpreted as a cut in a pre-defined graph connecting then pixels. Training the network should result in improved cuts, but no attempt to use a solver to find an optimal cut is made.

Here, we are considering a different setup, in which each new output of the neural network gives rise to a new instance of a combinatorial problem. A combinatorial algorithm is then used to find the optimal solution to the problem defined by the output, and the value of the objective function of the optimal solution is used as a loss. 
After each gradient update, the network will produce a new combinatorial problem instance, even for the same input sample. Iteratively, the network is expected to learn to produce combinatorial problem instances that have low optimal objective function value. For example, in sequence-to-sequence modeling, the network will output a new sentence that is supposed to closely match the desired sentence, leading to a new optimal sequence alignment problem to be solved. Initially, the optimal alignment will be poor, but as the network improves and the quality of the output sentences get higher, the optimal alignment scores will be lower.

Recently, progress in integrating combinatorial problems into differentiable models have been made by modifying combinatorial algorithms to use only differentiable elements \cite{tschiatschek2018differentiable,mensch2018differentiable,Chang_2019_CVPR}, for example smoothed max instead of max in dynamic programming. Another approach involves executing two runs of a non-differentiable, black-box combinatorial algorithm and uses the two solutions to define a differentiable interpolation \cite{poganvcic2019differentiation,rolinek2020deep}. Finally, differentiable linear programming and quadratic programming layers have been proposed recently \cite{amos2017optnet,agrawal2019differentiable,wilder2019melding,ferber2019mipaal}. 

The approaches above allow for differentiating through optimal solution vectors. In many cases, we are interested only in the optimal objective value, not the solution vector, and the approaches above introduce unnecessary overhead. We propose an approach for gradient-descent based training of a network $f(x;\beta)$ for supervised learning problems involving samples $(x,y)$ with the objective criterion of the form $L(f(x;\beta),y) + \mathrm{OptSolutionObjectiveValue}(P(f(x;\beta),y))$, where $L$ is a loss function, and $P$ is a combinatorial problem defined by the output of the network for feature vector $x$ and the true label $y$. 
We show that a broad class of combinatorial problems can be integrated into models trained using variants of gradient descent. Specifically, we show that for an efficiently solvable combinatorial problem that can be efficiently  expressed as an integer linear program, generalized gradients of the problem's objective value with respect to real-valued parameters defining the problem exist and can be efficiently computed from a single run of a black-box combinatorial algorithm. 
Using the above result, we show how generalized gradients of combinatorial problems can provide sentence-level loss for text summarization using differentiable encoder-decoder models that involve softmax or Gumbel softmax \cite{jang2016categorical}, and an multi-element loss for training classification models when only weakly supervised, bagged training data is available.

\section{Differentiable Combinatorial Losses}

\subsection{Background on Generalized Gradients}
A function $f: \Xset \rightarrow \Rset$ defined over a convex, bounded open set $\Xset \in \Rset^p$  is  Lipschitz continuous on an open set $B \in \Xset$ if there is a finite $K \in \Rset$ such that 
$\forall x,y \in B \;\; |f(x) - f(y)| \leq K \norm{x-y} $. A function is locally Lipschitz-continuous if for every point $x_0$ in its domain, there is a neighborhood $B_0$, an open ball centered at $x_0$, on which the function is Lipschitz-continuous. For such functions, a generalized gradient can be defined.
\begin{mydef}{\em\cite{clarke1975generalized}}
	Let $f: \Xset \rightarrow \Rset$ be Lipschitz-continuous in the neighborhood of $x \in \Xset$. Then, the {\em Clarke subdifferential} $\partial f(x)$ of $f$ at $x$ is defined as 
	\[
	\partial f(x) = \hull \left\{  \lim_{x_k \rightarrow x} \nabla f(x_k) \right\},
	\]
	where the limit is over all convergent sequences involving those $x_k$ for which gradient exists, and $\hull$ denotes convex hull, that is, the smallest polyhedron that contains all vectors from a given set. Each element of the set $\partial f(x)$ is called a {\em generalized gradient} of $f$ at $x$. 
\end{mydef}
The Rademacher theorem (see e.g. \cite{evans1992measure}) states that for any locally Lipschitz-continuous function the gradient exists almost everywhere; convergent sequences can be found.
In optimization algorithms, generalized gradients can be used in the same way as subgradients  \cite{redding1992learning}, that is, nondifferentiability may affect convergence in certain cases.

\subsection{Gradient Descent over Combinatorial Optimization}

Many combinatorial problems have linear objective function and can be intuitively expressed as integer linear programs (ILP), that is, linear programs with additional constraint that the solution vector involves only integers. Any ILP can be reduced to a linear program. 
Consider an ILP
\begin{align*}
\lpopt{z} = ILP(c,A',b') \; :=\; \mathrm{min}_{u} \;\;& c^T u \;\; \; \mathrm{s.t.} \;\;  A' u = b', \;\; u \geq 0,  \;\;  u \in \Zset^p, \nonumber
\end{align*}
with an optimal solution vector $\lpopt{u}$ and optimal objective value $\lpopt{z}$. Then, there exists a corresponding linear program $LP(c,A,b)$
\begin{align*}
\lpopt{z} = LP(c,A,b) \; := \; \mathrm{min}_{u} \;\;& c^T u \;\;\;  \mathrm{s.t.} \;\;  Au = b, \;\; u \geq 0,  \nonumber
\end{align*}
called {\em ideal  formulation} \cite{wolsey1989strong}, for which $\lpopt{u}$ is also an optimal solution vector, with the same objective value $\lpopt{z}$.
For a feasible, bounded $p$-dimensional integer program, we can view the pair  $(A',b')$ as a convex polyhedron $\setAp$, the set of all feasible solutions. Then, the pair  $(A,b)$ in the ideal formulation LP is defined as the set of constraints specifying the feasible set $\setA=\hull \left\{  \setAp \cap \Zset^p \right\}$. Convex hull of a subset of a convex set $\setAp$ cannot extend beyond $\setAp$, thus, $\setA$ is convex, contains all integer solutions from $\setAp$, and no other integer solutions.  The number of linear constraints in the ideal formulation may be exponential in $p$, and/or in $m$, the number of the original constraints in $\setAp$. Thus, the existence of the ideal formulation LP for an ILP may not have practical utility for solving the ILP. 

For a combinatorial problem and its corresponding ILP, we use the ideal formulation of the ILP as a conceptual tool to define generalized gradient of the objective value of the optimal solution to the combinatorial problem with respect to the parameters defining the combinatorial problem. Specifically, our approach first uses a single run of an efficient, black-box combinatorial algorithm to produce the optimal solution vector and the associated objective value. Then, the combinatorial problem is conceptually viewed as an instance of an ILP. A possibly exponentially large linear program (LP) equivalent to the ILP is then used, without actually being spelled out or solved, to derive generalized gradients based on the solution vector returned by the combinatorial algorithm.

First, we introduce several notions of efficiency of transforming a combinatorial problem into a linear integer program that will be convenient in defining the generalized gradients of combinatorial problems. 
\begin{mydef}
	Let $\Prob(\ProbIn)$ be a combinatorial problem that is parameterized by a continuous vector $\ProbIn \in \ProbSpace \subseteq \Rset^n$, where $\ProbSpace$ is simply connected and $n$ is the problem size, and let $k \in \Zset$ be a constant that may depend on the problem type but not on its size.
	Then, a combinatorial problem is 
	\begin{itemize}
		\item {\em primal-dual $\partial$-efficient} if it can be phrased as an  integer linear program involving $n$ variables, with $kn$ constraints in an LP formulation equivalent to the ILP, and the parameters $(A,b,c)$ of the LP formulation depend on $\ProbIn$ through (sub)differentiable  functions, $c=c(\ProbIn), A=A(\ProbIn), b=b(\ProbIn)$. 
		\item {\em primal  $\partial$-efficient} if it can be phrased as an integer linear program involving $n$ variables,  the parameters $\ProbIn$ of the problem influence the cost vector $c$ through a (sub)differentiable function $c=c(\ProbIn)$, and do not influence the constraints $A,b$.
		\item {\em dual  $\partial$-efficient} if it can be phrased as an integer linear program in which  the number of constraints in the equivalent LP formulation is $kn$, the parameters $\ProbIn$ of the problem influence $b$ through a (sub)differentiable function $b=b(\ProbIn)$, and do no influence the constraint matrix $A$ nor the cost vector $c$.
	\end{itemize}
	\label{def:eff}  
\end{mydef}
The class of $\partial$-efficient problems includes polynomially solvable combinatorial problems with objective function that is linear in terms of problem parameters. 
Typically, the functions $c=c(\ProbIn)$, $b=b(\ProbIn)$ and $A=A(\ProbIn)$ are either identity mapping or are constant; for example, in the LP for maximum network flow, the cost vector $c$ is composed directly of edge capacities, and $A$ an $b$ are constant for a given flow network topology, and do not depend on capacities.

For any polynomially solvable combinatorial problem, we can construct a $\poly(n)$-sized Boolean circuit for the algorithm solving it. For each $\poly(n)$-sized circuit, there is a linear program with $\poly(n)$ variables and constraints that gives the same solution (see \cite{dasgupta2008algorithms}, Chap. 7). For example, for MST in a graph with  $V$ vertices and $E$ edges, the Martin's ILP formulation \cite{martin1991using}  has only $\poly(V+E)$ constraints, but it is an extended formulation that involves $VE$ additional variables on top of the typical $E$ variables used in the standard ILP formulations for MST. Thus, we cannot use it to construct an ILP formulation that would make MST primal-dual $\partial$-efficient. Alternatively, there is an ILP for MST with one binary variable per edge, and the weight of the edge only influences the cost vector $c$, but to prohibit cycles in the solution there is a constraint for each cycle in the graph, thus the number of constraints  is not $\poly(n)$ for arbitrary graphs. These constraints are specified fully by the topology of the graph, not by the edge weights, so $w$ does not influence $A$ nor $b$, meeting the conditions for primal $\partial$-efficiency. The MST example shows that there are problems that are primal $\partial$-efficient and not primal-dual $\partial$-efficient.

Some polynomially solvable combinatorial problems  are not $\partial$-efficient in any of the above sense. For example, fixed-rank combinatorial problems with interaction costs \cite{lendl2019combinatorial} can be phrased succinctly as a bilinear program, but lead to prohibitively large linear programs both in terms of the number of variables and the number of constraints. 

For $\partial$-efficient problems, we can efficiently obtain generalized gradients of the objective value.  
\begin{theorem} \label{thm:ours}
	Consider a combinatorial problem $\Prob(\ProbIn)$ of size $n$, a parameter vector $\ProbIn$ from the interior of the parameter domain $\ProbSpace$, and an algorithm $\Alg(\ProbIn)$ for solving it in time $\poly(n)$. Let $\lpopt{z}$ be the optimal objective value returned by $\Alg$. Then,
	\begin{itemize}
		\item if $\Prob$ is primal  $\partial$-efficient, then the generalized gradients $\partial \lpopt{z}(\ProbIn)$ exist, and can be efficiently computed from $\lpopt{U}$, the set of primal solution of the ideal formulation of integer program corresponding to $\Prob$; 
		\item if $\Prob$ is dual  $\partial$-efficient, then the generalized gradients of $\partial \lpopt{z}(\ProbIn)$ exist, and  can be efficiently computed from $\lpopt{V}$, the set of all dual solution to the ideal formulation of the integer program corresponding to $\Prob$; 
		\item if $\Prob$ is primal-dual  $\partial$-efficient, then the generalized gradients of $A$ over $\ProbIn$ exist, and can be efficiently computed from $\lpopt{U}$ and $\lpopt{V}$, as defined above. 
	\end{itemize}
	\begin{proof}
		A series of results  \cite{gal1975rim,freund1985postoptimal,deWolf2000} shows that 
		if the optimal objective value $\lpopt{z}=LP(c,A,b)$ for a linear program is finite at $(c,A,b)$ and in some neighborhood of $(c,A,b)$, then 
		generalized gradients of $\lpopt{z}$ with respect to $c$, $b$, and $A$ exist and are 
		\begin{align*}
		\partial \lpopt{z}(c) = \lpopt{U},\;\;
		\partial \lpopt{z}(b) = \lpopt{V}, \;\;
		\partial \lpopt{z}(A)&= \left\{ - vu^T : (u,v) \in \lpopt{V} \times \lpopt{U} \right\}.
		\end{align*}
		We build on these results to obtain generalized gradients of the linear program corresponding to the combinatorial problem.
		For the first case in the theorem, definition \ref{def:eff} states that in the  linear program corresponding to $\Prob$, only the cost vector $c$ depends on $\ProbIn$, through a (sub)differentiable function $c=c(\ProbIn)$. Since $\ProbIn$ is in the interior of the parameter domain $\ProbSpace$, the objective value is finite over some neighborhood of $\ProbIn$.
		Then, 
		\begin{align*}
		\partial \lpopt{z}(\ProbIn) = \partial \lpopt{z}(c) \frac{\partial c}{\partial \ProbIn} = \frac{\partial c}{\partial \ProbIn}\lpopt{U} ,
		\end{align*}
		where the generalized gradient $\lpopt{z}(c)$ exists and is equal to $\lpopt{U}$.   
		\\ For the second case, the ideal formulation LP exists. Then, from definition \ref{def:eff} we have that 
		\begin{align*}
		\partial \lpopt{z}(\ProbIn) = \partial \lpopt{z}(b) \frac{\partial b}{\partial \ProbIn} = \frac{\partial b}{\partial \ProbIn}\lpopt{V}.
		\end{align*}
		\\ The third case is a direct extension of the first two cases. 
	\end{proof}
\end{theorem}

Theorem \ref{thm:ours} indicates that black-box combinatorial algorithms can be used to expand the range of transformations that can be efficiently utilized in neural networks. One immediate area of application is using them to specify a loss function. Consider a network $F(x;\beta)$ parameterized by a vector of tunable parameters $\beta$. The network transforms a batch of input samples $x$ into a batch of outputs $y=F(x;\beta)$. Then, in the broadest primal-dual  $\partial$-efficient case, $y$ is used, possibly with $x$ , to formulate parameters $(c,A,b)=g(x,y)$ of a linear program corresponding to the combinatorial problem, through some (sub)differentiable function $g$. 
For a given $\beta$ and given batch $(x,y)$, we can then define loss as a function of the optimal objective value of the linear program corresponding to the combinatorial problem resulting from $g(x,F(x;\beta))$, $L(\beta)=h(\lpopt{z}(c,A,b))$. This approach, summarized in Algorithm 1, allows us to obtain the generalized gradient of the loss with respect to $\beta$ as long as functions $g$ and $h$ are differentiable. For clarity, in Algorithm 1, we did not consider functions $h$ depending not just on $z$ but also on $x$ or $y$, but the extension is straightforward.
\begin{algorithm}[t]
	\renewcommand{\algorithmicrequire}{\textbf{Input:}}
	\renewcommand{\algorithmicensure}{\textbf{Output:}}
	\caption{Minimization of a combinatorial loss}
	\label{genericLPLossScheme}
	\begin{algorithmic}[1]
		\Require batch $x \subset \Xset$, $y \subset \Yset$, network $F(x;\beta)$,  functions $g, h$, combinatorial algorithm $\Alg$
		\Ensure Loss and its generalized gradient, $L(\beta),\partial L(\beta)$ 
		\Procedure{  CombLossMin($x,y,\beta,F,g,h,\Alg$) }{}
		\State forward pass $y=F(x;\beta)$
		\State forward pass $(c,A,b)=g(x,y)$
		\State run combinatorial solver to find optimal objective value $\lpopt{z}=\Alg(c,A,b)$  and optimal $\;\;\;\;\;\;\;\;$ \mbox{ } $\;\;\;\;\;\;\;\;$ primal  and/or dual solution vectors $\lpopt{u}$, $\lpopt{v}$
		\State forward pass $L(\beta)=h(\lpopt{z})$
		\State backward pass through $h$: $\partial L/ \partial \lpopt{z}$
		\State backward pass through $\Alg$: $\partial \lpopt{z}(c)=\lpopt{u}$,  $\partial \lpopt{z}(b)=\lpopt{v}$, $\partial \lpopt{z}(A)=- \lpopt{v}\lpopt{u}$
		\State backward pass through $g$ and $F$
		\State $\partial L(\beta)=\frac{\partial L}{\partial z} \Big( \lpopt{u} \frac{\partial c}{\partial \beta} - \lpopt{v}\lpopt{u} ^T\frac{\partial A}{\partial \beta} + \lpopt{v}\frac{\partial c}{\partial \beta}\Big)$
		\State \Return $L(\beta)$,  $\partial L(\beta)$
		\EndProcedure
	\end{algorithmic}
\end{algorithm}

\section{Example Use Cases and Experimental Validation}

\subsection{Differentiating over Bipartite Matching for Weakly-supervised Learning}

To illustrate gradient descent over a combinatorial loss, we first focus on a simple image recognition problem. 
Consider a photo of a group of people with a caption listing each of the persons in the picture, but missing the "from left to right" part. Given a collection of such labeled photos, can a model learn to recognize individual faces? Similarly, consider a shopping cart and a printout from the register. Given a collection of unordered shopping carts together with matching receipts, can a model learn to recognize individual shopping items? These are example of a weakly-supervised learning where the goal is to learn to classify previously unseen feature vectors, but a training sample is a bag of feature vectors accompanied by a bag of correct labels, instead of a feature-vector and a correct label.  We are not told which class belongs to which sample, which prevents us from directly using the standard cross-entropy loss. 

More formally, consider a $d$-class classification problem, and a network $F(x_i;\beta)$ that for sample $x_i$ returns a $d$-dimensional vector of class probabilities, $p_i$, with $p_{i}$ denoting the predicted conditional probability of class $j$ given feature vector $x_i$. Let $y_i$ denote a $d$-dimensional, one-hot representation of the true class label of sample $x_i$. 

In weakly supervised learning involving bags of size $b$, we are given a tuple of $b$ feature vectors,  $X=\left(x_j\right)_{j=1}^b$, and a tuple of permuted labels  $Y=\left(y_{\sigma(i)}\right)_{i=1}^b$ as one-hot-vectors, for some  permutation $\sigma$; we will refer to the $j$-th element of the tuple $Y$ as $Y_j$. The permutation $\sigma$ is unknown, thus using a loss $\ell(p_j, Y_j)=\ell(p_j, y_{\sigma(i)})$ makes no sense, since most likely $ i \neq j$. 
While the permutation is unknown, with repeated presentation of bags of samples and bags of corresponding labels, we do have some information connecting the feature vector to classes. Intuitively, we can try to match feature vectors in the bag to the class labels using the information in the model’s probability distribution, that is, find   permutation $\hat{\sigma}$ optimal in the average loss sense $\min_{\hat{\sigma}} \sum_{j=1}^b
\ell(p_j, \hat{\sigma}(Y)_j)$.
If the class conditional probabilities $p_j$ resulting from the model perfectly match the one-hot vectors, the optimal  $\hat{\sigma}$ will be the inverse of the permutation $\sigma$. 

A $b$-element permutation can be represented by a $b \times b$ permutation matrix $M$. To find $M$, we define with a $b \times b$ matrix $C$ with $C_{jk}=\ell(p_j,Y_k)$ -- the elements $C_{jk}$ correspond to edge weight in a bipartite graph with feature vectors on one side, and labels on the other side. 
We use a combinatorial solver, for example the Hungarian method with computational complexity $\BigOh{b^3}$, to find the the permutation matrix $\lpopt{C} = \arg\min_C \inner{C}{M}_F$ minimizing the Frobenius inner product of $C$ and $M$.

\begin{figure}[t]
	\centering
	\includegraphics[width=0.49\linewidth]{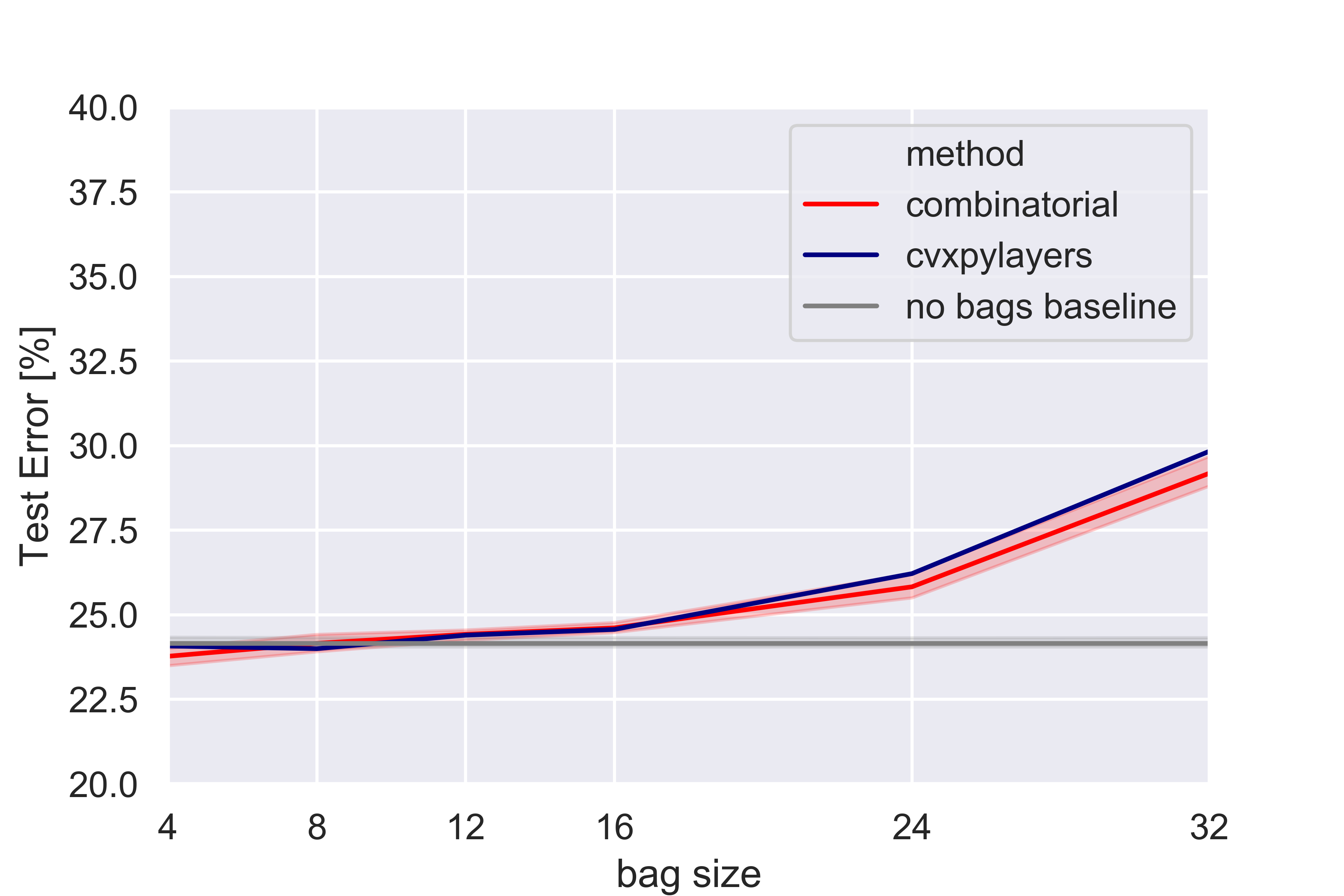} 
	\includegraphics[width=0.49\linewidth]{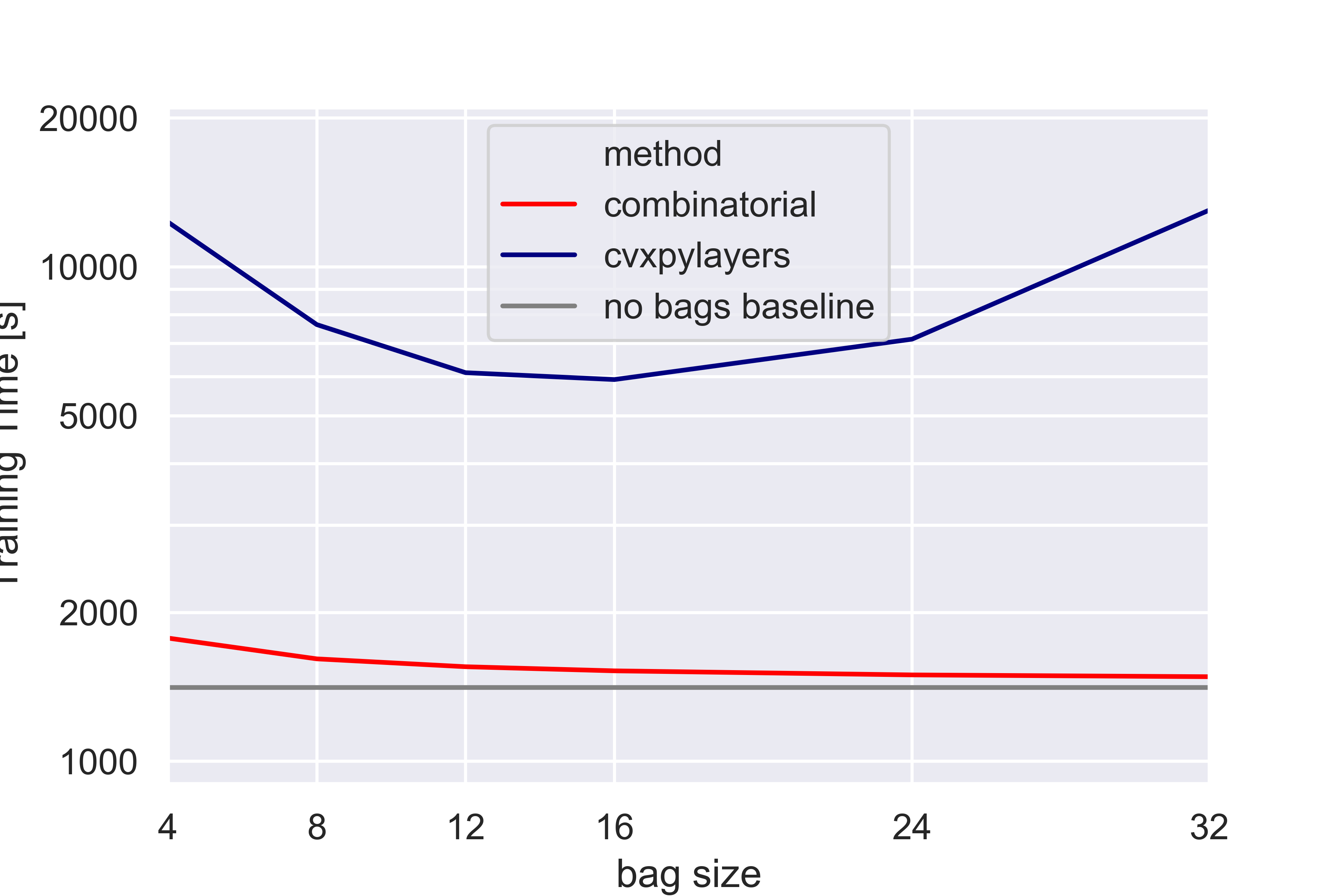} 
	\caption{Test set error (left) and total training time (right) for a classifier trained using the proposed combinatorial loss and, for comparison, a loss based on cvxpylayers \cite{agrawal2019differentiable}, for increasing bag sizes. A supervised model with true label available for each individual sample, which corresponds to bag of size one, is used as a baseline lower bound on the error that the bag-trained models should attempt to match. Mean, and the  95\% confidence interval of the mean, are shown. \label{fig:eBags}}
\end{figure}

To test the approach, we used the  CIFAR100 benchmark image dataset. We trained 5 independent baseline supervised models with ResNet18 architecture \cite{zagoruyko2016wide} (see Supplementary Material for details), that is, models where each image is a separate sample with its true class available for loss calculation. We used cross-entropy loss $\ell(p,y)=-\inner{\log p}{y}$, where the logarithm is applied element-wise. To evaluate the combinatorial loss, during training we explored image bags of samples consisting of  4, 8, 12, 16, 24, or 32 images, and correct but shuffled image labels, and trained 5 independent models for each bag size with the combinatorial loss based on weighted bipartite graph matching, using cross-entropy as the loss defining the edge weights $C_{jk}$. To avoid situations where the combinatorial loss is aided by bags with mostly one class, we ignored any bag that has less than 75\% of different classes, that is, for bag of size 8, we only consider bags that consist of at least 6 different classes. During testing, same as in the baseline model experiments, each image had the matching label available for test error calculations. 
For comparison, we trained a model with the same setup of image bags using cvxpylayers \cite{agrawal2019differentiable}, a recently proposed methods for differentiable layers defined by conic programs. In contrast to our approach, which uses a combinatorial algorithm and relies on the LP formulation of the weighted bipartite matching only for the definition of gradients, cvxpylayers solve the linear program in order to obtain gradients.

Test error for CIFAR100 of the training set reshuffled into bags after each epoch (Fig. \ref{fig:eBags}, left) shows that for bag sizes up to twelve elements, weak supervision through weighted bipartite graph matching is almost as effective as supervised learning with true label available for each individual image, that is, bag of size one. Training using the proposed combinatorial loss and using cvxpylayers leads to very similar error rates (Fig. \ref{fig:eBags}, left), but is much faster  (Fig. \ref{fig:eBags}, right); especially for larger bag sizes, the combinatorial loss introduces negligible overhead over ResNet18 computations, while cvxpylayers result in five-to-ten-fold increase in the training time, depending on the bag size. 
These results show that the generalized gradient over combinatorial optimization is effective in providing training signal to train a large neural network, and can do it much faster than the state-of-the-art existing approach.

\subsection{Differentiating over Global Sequence Alignment  for Sentence-level Loss in Sequence-to-Sequence Models}

Another use case where a combinatorial loss is advantageous occurs in to sequence-to-sequence natural language models. We used a standard encoder-decoder architecture for the model (see Supplementary Material for details).  The encoder takes the source sequence on input and prepares a context vector capturing the source sequence. The decoder is a recurrent network that outputs the predicted sequence one token at a time, based on the context vector and the output of the previous step. The output of the decoder at a step $t$ is a vector of probabilities $p_t$ over the set of all possible output tokens. 

\begin{figure}[!t]
	\centering
	\includegraphics[width=0.95\linewidth]{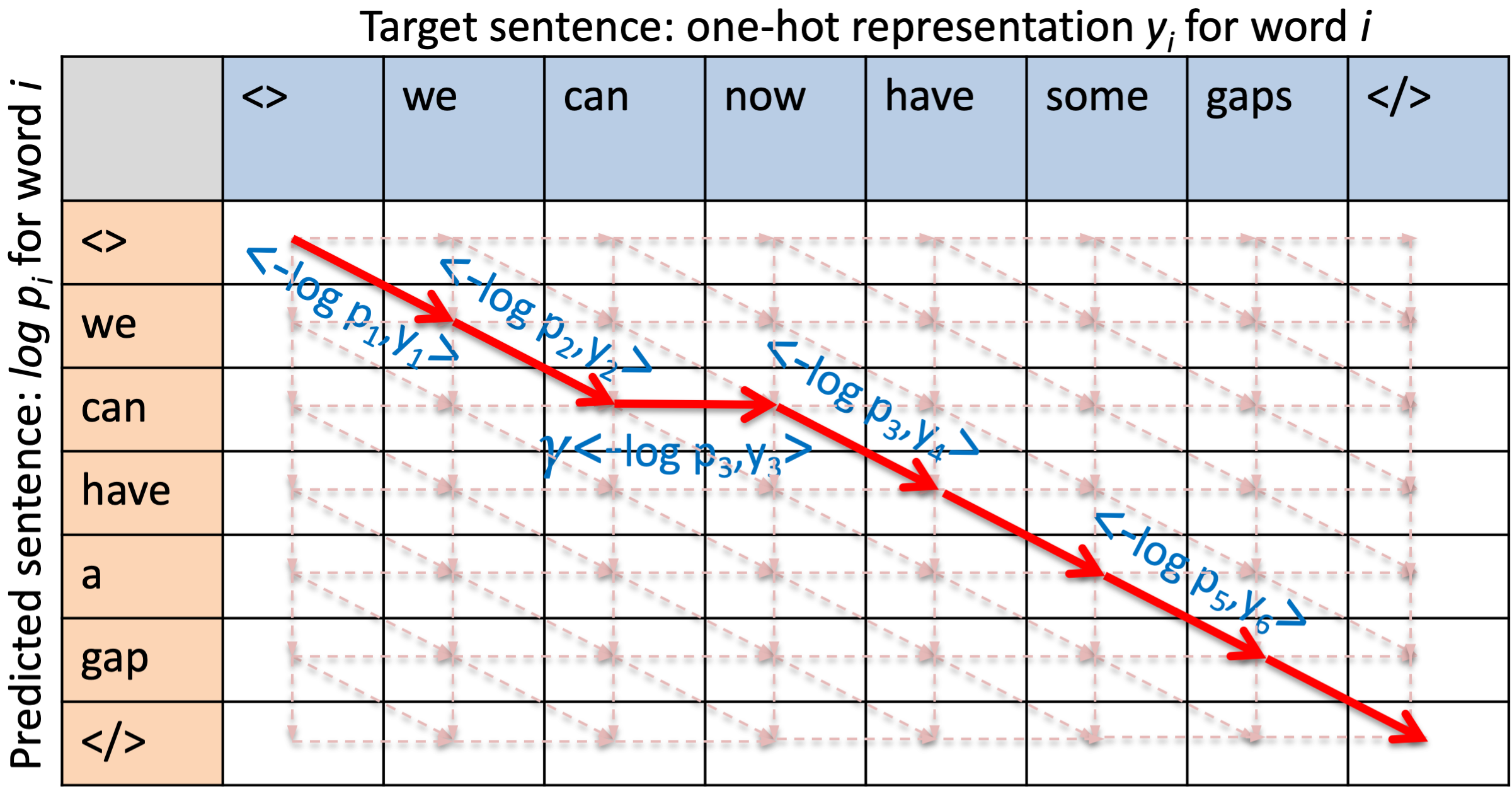}
	\caption{A directed acyclic graph (DAG) corresponding to the  global sequence alignment between the target sequence and the sequence predicted by the RNN model. Each node, except the end of sequence indicator $</>$, has out-degree of three: a diagonal edge corresponding to a match between the predicted and the target sequence, a horizontal edge corresponding to a gap in the predicted sequence, and a vertical edge corresponding to a gap in the target sequence. Optimal sequence alignment is depicted in red, with the weights -- the alignment costs -- of the selected edges in blue.\label{fig:GSA}}
\end{figure}

Existing encoder-decoder models use cross-entropy loss to compare predicted probabilities $p_t$ to the target word at position $t$, encoded as one-hot vector $y_t$. Instead of a sequence-level optimization, position-specific cross entropy loss results in an averaged token-level optimization. We hypothesize this has detrimental effect on the training process of differentiable sequence-to-sequence models that involve softmax or Gumbel-softmax \cite{jang2016categorical} as the mechanism for feeding the output of the previous step of the decoder as input for the next step. For example, a recurrent model that learned to output almost all of the target sentence correctly but is still making the mistake of missing one word early in the sentence will have very high loss at all the words following the missing word -- correcting the mistake should involve keeping most of the model and focusing on the missing word, but with position-specific loss, all the outputs are considered wrong and in need of correction.

Gaps or spurious words in the output sequence can be treated naturally if we consider global sequence alignment (GSA) as the loss. Global sequence alignment \cite{needleman1970general} is a combinatorial problem in which two sequences are aligned by choosing, at each position, to either match a token from one sequence to a token from the other, or to introduce a gap in one or the other sequence; each choice has a cost (see Fig. \ref{fig:GSA}). 
In sequence-to-sequence modeling, the cost of matching the decoder's output from position $i$ to the target sequence token as position $k$ will be given by $\inner{-\log p_i}{y_k}$. The cost of a gap, that is, of a horizontal or a vertical move in Fig. \ref{fig:GSA}, is specified in a way that promotes closing of the gap; we use the cost of diagonal move from that position as the cost of the gap, multiplied by a scalar $\gamma > 1$ to prioritize closing the gaps over improving the matchings. In our experiments, we used $\gamma=1.5$. 
The GSA problem can 
stated as a linear program with $p$ variables and $m+1$ constraints, with the costs of the moves forming the right-hand side of the constraints. Thus, by Theorem \ref{thm:ours}, the generalized gradient of the minimum global sequence alignment with respect to matching and gap costs is efficiently available. 

In experiments involving global sequence alignment in sequence-to-sequence models, we used an encoder-decoder sequence-to-sequence architecture with bidirectional forward-backward RNN encoder and an attention-based RNN decoder \cite{luong2015effective}, as implemented in PyTorch-Texar \cite{hu2018texar}. While this architecture is no longer the top performer in terms of ROUGE metric -- currently, large pre-trained self-attention models are the state-of-the-art 
-- it is much more efficient in training, allowing for experimenting with different loss functions. In evaluating the combinatorial GSA loss, we used text summarization task involving the GIGAWORD dataset  \cite{graff2003english} as an example of a sequence-to-sequence problem. We used test set ROUGE 1, 2, and L scores \cite{lin2004rouge} as the measure of quality of the summarizations.

\begin{table*}[t]
	\caption{Results for the GIGAWORD text summarization task using ROUGE-1, ROUGE-2, and ROUGE-L metrics. For MLE and our combinatorial method (GSA-L), results are given as mean(std.dev.) over five independent runs with different random seed. For the method involving cvxpylayers (GSA-C) we only performed one run. We report test set values for the epoch that minimizes the total ROUGE score on a separate validation set. Time values are reported per epoch. \label{tab:ROUGE}}
	\begin{center}
		\begin{tabular}{lllllll}
			\textbf{Loss Type}  & \textbf{ROUGE-Total} &	\textbf{ROUGE-1}  &\textbf{ROUGE-2} &\textbf{ROUGE-L} &\textbf{Epoch} & \textbf{Time} \\
			\hline 
			\multicolumn{7}{c}{\bf Softmax} \\ 
			MLE  & 72.80(0.38) &32.45(0.15) &	11.95(0.22) &	28.39(0.20) &	18.4(1.5) & 8 min\\
			GSA - C & 32.18 &17.04 &	2.49  &	12.65 &	3 & 9 hr \\
			GSA - L & 76.36(0.60)  &34.05(0.21) &	12.31(0.20)  &	29.99(0.24) &	15.4(2.5) & 17 min\\ \hline
			
			\multicolumn{7}{c}{\bf Gumbel-softmax} \\ 
			MLE  & 67.50(0.20) &31.25(0.18) &	9.72(0.26) &	26.52(0.08) &	18.0(2.8) & 9 min \\
			GSA - L &  72.62(0.51) &33.25(0.15) &	10.60(0.22) &	28.77(0.17) &	14.0(1.9) & 17min\\ \hline
		\end{tabular}
	\end{center}
\end{table*}

The results in Table \ref{tab:ROUGE} show that the combinatorial loss based on the global sequence alignment leads to improved text summarization results in all three ROUGE metrics compared to position-specific maximum likelihood training, both for the softmax and the Gumbel-softmax approach. The increase in accuracy comes at the cost of doubling the training time. 
The proposed combinatorial approach is much more accurate and efficient than the recently proposed cvxpylayers method. The running time for the cvxpylayers approach is orders of magnitude slower. The cvxpylayers solver managed to reduce the training loss for several initial epochs, after which solver errors start to occur and the learning process diverges. In order to confirm this behavior, we performed 3 additional runs of the cvxpylayers-based training for the softmax model. In all cases, the loss dropped from the initial value in the 90-95 range to above 50, after which it increased to  500 or more. For comparison, the proposed combinatorial approach and the MLE approach reach loss in the 30-32 range by epoch 10. 

\section{Related Work}

Recently, \cite{tschiatschek2018differentiable} proposed an approximate solver for submodular function maximization that uses differentiable elements and allows for differentiating through the solver. Differentiable solvers are also considered in \cite{mensch2018differentiable}, where the maximum operation in dynamic programming  is replaced by smoothed max. Similar approach is used in  differentiable dynamic time warping \cite{Chang_2019_CVPR}. 
Several authors used a differential approximation to linear program solutions instead of introducing differentiable operations into combinatorial algorithms. WGAN-TS \cite{liu2018two} solves an LP to obtain the exact empirical Wasserstein distance. Then, to circumvent lack of differentiability of linear programs, WGAN-TS proceeds by training a neural network to approximate the LP solution in order to obtain gradients. In seq2seq-OT \cite{chen2019improving}, an approximation is used to model optimal transport between word embeddings serving as a regularizer in training sequence-to-sequence models.  These approximation approaches are limited to specific problems and preclude using off-the-shelf combinatorial solvers.

Recently, an approach that performs two runs of a non-differentiable black-box combinatorial algorithm and uses the two solutions to define a differentiable interpolation \cite{poganvcic2019differentiation,rolinek2020deep}. The approach allows for using off-the-shelf combinatorial solvers  but, by requiring one run for the forward phase, and a second run for a slightly perturbed problem for the backward phase, doubles the time overhead compared to our approach.

An alternative approach is to use mathematical programming solvers in gradient-trained neural networks.
OptNet \cite{amos2017optnet} provides differentiable quadratic programming layers, and an efficient GPU-based batch solver, qpth. QP layer can be used not only for QP problems, for also for stochastic programming solved via sequential quadratic programming \cite{donti2017task}. Cvxpylayers \cite{agrawal2019differentiable} generalizes this approach  to a broad class of convex optimization problems expressed as cone programs, which include QP and LP as special cases, using conic solver based on ADMM, providing a general-purpose package based on the easy-to-use interface of cvxpy, with speed comparable to qpth for QP problems. Other authors \cite{wilder2019melding,ferber2019mipaal} focus on LP problems,  regularize them by adding the quadratic term, and use a QP solver as in OptNet  to obtain the solution and its gradient. Quadratic smoothing is also used in \cite{djolonga2017differentiable} in submodular set function minimization.
Compared to these methods, in the approach proposed here, linear programming is used only as a theoretical tool that allows for defining a mapping from the solution to a combinatorial problem to its gradient. The solution is obtained by a single run of a combinatorial algorithm, which, as our experiments confirm, is faster than using mathematical programming and not affected by numerical instability and convergence problems. 

\section*{Acknowledgments}
T.A. is supported by NSF grant IIS-1453658.

\bibliographystyle{alpha}

\begin{thebibliography}{ZJRP{\etalchar{+}}15}
	
	\bibitem[AAB{\etalchar{+}}19]{agrawal2019differentiable}
	Akshay Agrawal, Brandon Amos, Shane Barratt, Stephen Boyd, Steven Diamond, and
	J~Zico Kolter.
	\newblock Differentiable convex optimization layers.
	\newblock In {\em Advances in Neural Information Processing Systems}, pages
	9558--9570, 2019.
	
	\bibitem[AK17]{amos2017optnet}
	Brandon Amos and J~Zico Kolter.
	\newblock Optnet: Differentiable optimization as a layer in neural networks.
	\newblock In {\em Proceedings of the 34th International Conference on Machine
		Learning}, pages 136--145, 2017.
	
	\bibitem[BVZ01]{boykov2001fast}
	Yuri Boykov, Olga Veksler, and Ramin Zabih.
	\newblock Fast approximate energy minimization via graph cuts.
	\newblock {\em IEEE Transactions on Pattern Analysis and Machine Intelligence},
	23(11):1222--1239, 2001.
	
	\bibitem[CHS{\etalchar{+}}19]{Chang_2019_CVPR}
	Chien-Yi Chang, De-An Huang, Yanan Sui, Li~Fei-Fei, and Juan~Carlos Niebles.
	\newblock D3tw: Discriminative differentiable dynamic time warping for weakly
	supervised action alignment and segmentation.
	\newblock In {\em The IEEE Conference on Computer Vision and Pattern
		Recognition (CVPR)}, June 2019.
	
	\bibitem[Cla75]{clarke1975generalized}
	Frank~H Clarke.
	\newblock Generalized gradients and applications.
	\newblock {\em Transactions of the American Mathematical Society},
	205:247--262, 1975.
	
	\bibitem[CLRS09]{Cormen}
	Thomas~H. Cormen, Charles~E. Leiserson, Ronald~L. Rivest, and Clifford Stein.
	\newblock {\em Introduction to Algorithms, Third Edition}.
	\newblock The MIT Press, 3rd edition, 2009.
	
	\bibitem[CZZ{\etalchar{+}}19]{chen2019improving}
	Liqun Chen, Yizhe Zhang, Ruiyi Zhang, Chenyang Tao, Zhe Gan, Haichao Zhang, Bai
	Li, Dinghan Shen, Changyou Chen, and Lawrence Carin.
	\newblock Improving sequence-to-sequence learning via optimal transport.
	\newblock In {\em International Conference on Learning Representations}, page
	arXiv:1901.06283, 2019.
	
	\bibitem[DAK17]{donti2017task}
	Priya Donti, Brandon Amos, and J~Zico Kolter.
	\newblock Task-based end-to-end model learning in stochastic optimization.
	\newblock In {\em Advances in Neural Information Processing Systems}, pages
	5484--5494, 2017.
	
	\bibitem[DK17]{djolonga2017differentiable}
	Josip Djolonga and Andreas Krause.
	\newblock Differentiable learning of submodular models.
	\newblock In {\em Advances in Neural Information Processing Systems}, pages
	1013--1023, 2017.
	
	\bibitem[DPV08]{dasgupta2008algorithms}
	Sanjoy Dasgupta, Christos~H Papadimitriou, and Umesh~Virkumar Vazirani.
	\newblock {\em Algorithms}.
	\newblock McGraw-Hill Higher Education, 2008.
	
	\bibitem[DWS00]{deWolf2000}
	Daniel De~Wolf and Yves Smeers.
	\newblock Generalized derivatives of the optimal value of a linear program with
	respect to matrix coefficients.
	\newblock In {\em Technical report.} Universit{\'e} Catholique de Louvain,
	2000.
	
	\bibitem[Eva92]{evans1992measure}
	LawrenceCraig Evans.
	\newblock {\em Measure theory and fine properties of functions}.
	\newblock Routledge, 1992.
	
	\bibitem[Fre85]{freund1985postoptimal}
	Robert~M Freund.
	\newblock Postoptimal analysis of a linear program under simultaneous changes
	in matrix coefficients.
	\newblock In {\em Mathematical Programming Essays in Honor of George B. Dantzig
		Part I}, pages 1--13. Springer, 1985.
	
	\bibitem[FWDT19]{ferber2019mipaal}
	Aaron Ferber, Bryan Wilder, Bistra Dilina, and Milind Tambe.
	\newblock {MIPaaL}: Mixed integer program as a layer.
	\newblock {\em arXiv preprint arXiv:1907.05912}, 2019.
	
	\bibitem[Gal75]{gal1975rim}
	Tomas Gal.
	\newblock Rim multiparametric linear programming.
	\newblock {\em Management Science}, 21(5):567--575, 1975.
	
	\bibitem[GC03]{graff2003english}
	David Graff and C~Cieri.
	\newblock English gigaword corpus.
	\newblock {\em Linguistic Data Consortium}, 2003.
	
	\bibitem[HST{\etalchar{+}}18]{hu2018texar}
	Zhiting Hu, Haoran Shi, Bowen Tan, Wentao Wang, Zichao Yang, Tiancheng Zhao,
	Junxian He, Lianhui Qin, Di~Wang, et~al.
	\newblock Texar: A modularized, versatile, and extensible toolkit for text
	generation.
	\newblock {\em arXiv preprint arXiv:1809.00794}, 2018.
	
	\bibitem[JGP16]{jang2016categorical}
	Eric Jang, Shixiang Gu, and Ben Poole.
	\newblock Categorical reparameterization with {Gumbel-softmax}.
	\newblock In {\em International Conference on Learning Representations ICLR'17.
		arXiv:1611.01144}, 2016.
	
	\bibitem[KHL16]{kusner2016gans}
	Matt~J Kusner and Jos{\'e}~Miguel Hern{\'a}ndez-Lobato.
	\newblock Gans for sequences of discrete elements with the gumbel-softmax
	distribution.
	\newblock {\em arXiv preprint arXiv:1611.04051}, 2016.
	
	\bibitem[L{\'C}P19]{lendl2019combinatorial}
	Stefan Lendl, Ante {\'C}usti{\'c}, and Abraham~P Punnen.
	\newblock Combinatorial optimization with interaction costs: Complexity and
	solvable cases.
	\newblock {\em Discrete Optimization}, 33:101--117, 2019.
	
	\bibitem[LDJ{\etalchar{+}}16]{lin2016scribblesup}
	Di~Lin, Jifeng Dai, Jiaya Jia, Kaiming He, and Jian Sun.
	\newblock Scribblesup: Scribble-supervised convolutional networks for semantic
	segmentation.
	\newblock In {\em Proceedings of the IEEE Conference on Computer Vision and
		Pattern Recognition}, pages 3159--3167, 2016.
	
	\bibitem[LGS18]{liu2018two}
	Huidong Liu, Xianfeng Gu, and Dimitris Samaras.
	\newblock A two-step computation of the exact {GAN Wasserstein} distance.
	\newblock In {\em International Conference on Machine Learning}, pages
	3165--3174, 2018.
	
	\bibitem[Lin04]{lin2004rouge}
	Chin-Yew Lin.
	\newblock Rouge: A package for automatic evaluation of summaries.
	\newblock In {\em Text summarization branches out}, pages 74--81, 2004.
	
	\bibitem[LPM15]{luong2015effective}
	Minh-Thang Luong, Hieu Pham, and Christopher~D Manning.
	\newblock Effective approaches to attention-based neural machine translation.
	\newblock In {\em Proceedings of the 2015 Conference on Empirical Methods in
		Natural Language Processing}, pages 1412--–1421, 2015.
	
	\bibitem[Mar91]{martin1991using}
	R~Kipp Martin.
	\newblock Using separation algorithms to generate mixed integer model
	reformulations.
	\newblock {\em Operations Research Letters}, 10(3):119--128, 1991.
	
	\bibitem[MB18]{mensch2018differentiable}
	Arthur Mensch and Mathieu Blondel.
	\newblock Differentiable dynamic programming for structured prediction and
	attention.
	\newblock In {\em Proceedings of the 35th International Conference on Machine
		Learning}, pages 3462--3471, 2018.
	
	\bibitem[MTAB19]{marin2019beyond}
	Dmitrii Marin, Meng Tang, Ismail~Ben Ayed, and Yuri Boykov.
	\newblock Beyond gradient descent for regularized segmentation losses.
	\newblock In {\em IEEE Conference on Computer Vision and Pattern Recognition
		(CVPR)}, 2019.
	
	\bibitem[NW70]{needleman1970general}
	Saul~B Needleman and Christian~D Wunsch.
	\newblock A general method applicable to the search for similarities in the
	amino acid sequence of two proteins.
	\newblock {\em Journal of molecular biology}, 48(3):443--453, 1970.
	
	\bibitem[PPM{\etalchar{+}}20]{poganvcic2019differentiation}
	Marin~Vlastelica Pogan{\v{c}}i{\'c}, Anselm Paulus, Vit Musil, Georg Martius,
	and Michal Rolinek.
	\newblock Differentiation of blackbox combinatorial solvers.
	\newblock In {\em International Conference on Learning Representations}, 2020.
	
	\bibitem[RD92]{redding1992learning}
	Nicholas~J Redding and Tom Downs.
	\newblock Learning in feedforward networks with nonsmooth functions.
	\newblock In {\em Advances in Neural Information Processing Systems}, pages
	1056--1063, 1992.
	
	\bibitem[RSZ{\etalchar{+}}20]{rolinek2020deep}
	Michal Rol{\'\i}nek, Paul Swoboda, Dominik Zietlow, Anselm Paulus, V{\'\i}t
	Musil, and Georg Martius.
	\newblock Deep graph matching via blackbox differentiation of combinatorial
	solvers.
	\newblock {\em arXiv preprint arXiv:2003.11657}, 2020.
	
	\bibitem[Sch03]{schrijver2003combinatorial}
	Alexander Schrijver.
	\newblock {\em Combinatorial optimization: polyhedra and efficiency},
	volume~24.
	\newblock Springer Science \& Business Media, 2003.
	
	\bibitem[TSK18]{tschiatschek2018differentiable}
	Sebastian Tschiatschek, Aytunc Sahin, and Andreas Krause.
	\newblock Differentiable submodular maximization.
	\newblock In {\em Proceedings of the 27th International Joint Conference on
		Artificial Intelligence}, IJCAI’18, page 2731–2738. AAAI Press, 2018.
	
	\bibitem[WDT19]{wilder2019melding}
	Bryan Wilder, Bistra Dilkina, and Milind Tambe.
	\newblock Melding the data-decisions pipeline: Decision-focused learning for
	combinatorial optimization.
	\newblock In {\em The Thirty-Third Conference on Artificial Intelligence
		(AAAI)}, pages 1658--1665, 2019.
	
	\bibitem[Wol89]{wolsey1989strong}
	Laurence Wolsey.
	\newblock Strong formulations for mixed integer programming: a survey.
	\newblock {\em Mathematical Programming}, 45(1):173--191, 1989.
	
	\bibitem[ZJRP{\etalchar{+}}15]{zheng2015conditional}
	Shuai Zheng, Sadeep Jayasumana, Bernardino Romera-Paredes, Vibhav Vineet,
	Zhizhong Su, Dalong Du, Chang Huang, and Philip~HS Torr.
	\newblock Conditional random fields as recurrent neural networks.
	\newblock In {\em Proceedings of the IEEE International Conference on Computer
		Vision}, pages 1529--1537, 2015.
	
	\bibitem[ZK16]{zagoruyko2016wide}
	Sergey Zagoruyko and Nikos Komodakis.
	\newblock Wide residual networks.
	\newblock In {\em International Conference on Learning Representations ICLR'17.
		arXiv:1605.07146}, 2016.
	
\end{thebibliography}

\newcommand{\etalchar}[1]{$^{#1}$}
\newcommand{\noopsort}[1]{}

\appendix
\section{Details of the Experimental Setup}

\subsection{Differentiating over Bipartite Matching for Weakly-supervised Learning}
In experiments involving weakly-supervised learning for image recognition, we used the ResNet18 architecture \cite{zagoruyko2016wide} involving residual, convolutional blocks.

We used the  CIFAR100 benchmark image dataset, with the standard training-testing split, as the source of images and labels.
Training set batch size was set to 384 elements divided into bags, that is, for bag size 32, we are solving 12 weighted matching problems, one for each bag. We trained all the networks using a cyclic learning rate with a maximum of 0.1 for 80 epochs, following by another cycle with maximum learning rate of 0.05 for 40 epochs. 

\subsection{Differentiating over Global Sequence Alignment  for Sentence-level Loss in Sequence-to-Sequence Models}

In experiments involving global sequence alignment in sequence-to-sequence models, we used an encoder-decoder sequence-to-sequence architecture with bidirectional forward-backward RNN encoder and an attention-based RNN decoder \cite{luong2015effective}, as implemented in PyTorch-Texar \cite{hu2018texar}. 
In both the encoder and the decoder we used 256 units and dropout rate of 0.2. We used batch size of 128, and learning rate of 0.001 with Adam optimizer.
During inference, we used beam search. During training, to have a differentiable decoder, we use two alternative approaches. First, we feed the probabilities resulting from softmax layer applied to the outputs of the RNN directly. Second, inputs to the RNN are provided by the straight-through Gumbel-softmax distribution \cite{jang2016categorical}, which is an approximation of the categorical distribution from which one-hot, single-token outputs are sampled. 
In both softmax and Gumbel-softmax, we use annealing of the temperature parameter $\tau$. 
As it has been used previously \cite{kusner2016gans}, we start with a high value of the temperature parameter $\tau$  and reduced it as training progresses. Specifically, we started with $\tau=5$,  reduced it by 0.5 in each epoch until value of 1.0 is reached, and then kept it fixed at 1.0. 

In evaluating the combinatorial GSA loss, we used text summarization task involving the GIGAWORD dataset  \cite{graff2003english} as an example of a sequence-to-sequence problem. We used the same preprocessing as \cite{chen2019improving}, that is, we used 200K examples in training. We used the validation set to select the best model epoch, and reported results on a separate test set.

\end{document}